\documentclass{edm_article}
\usepackage[hidelinks]{hyperref}
\usepackage{needspace}
\usepackage{graphicx}
 \usepackage{float}   
 \usepackage{placeins}
 \usepackage{dblfloatfix}
 \usepackage{capt-of} 
 \usepackage[caption=false,font=footnotesize]{subfig} 
 \usepackage{cuted}

\begin{document}

\title{Learning to Grade Efficiently: A Bandit-Driven Prompt-Selection Framework for Low-Cost LLM Essay Scoring}

\numberofauthors{2}
\author{%
  \alignauthor Olga Manakina\\
      \affaddr{Carleton University}\\
      \email{olgamanakina@cmail.carleton.ca}
  \alignauthor Igor Bogdanov\\
      \affaddr{Carleton University}\\
      \email{igorbogdanov@cmail.carleton.ca}
}

\maketitle

\begin{abstract}
Large Language Models (LLMs) demonstrate strong capabilities in automated essay scoring (AES), but contemporary approaches typically employ fixed prompt selection, failing to address operational cost concerns and evolving optimal configurations. We propose a novel cost-aware approach that treats each prompt type as an arm in a multi-armed bandit (MAB) controller, enabling adaptive selection of optimal prompting strategies during inference. Our experiments on IELTS Writing Task 2 essays show that the MAB framework achieves comparable scoring accuracy to exhaustive grid search while reducing LLM calls by 78.4\% to find the best grading approach (Table~\ref{tab:detailed_resource_comparison}). We implemented four distinct grading recipes (multi-step vs. single-step assessment, with vs. without calibration examples) and found that the multi-step approach with examples achieves the highest accuracy. By tracking token usage and latency alongside agreement metrics, we produce the first cost-reliability learning curves for essay scoring, providing actionable insights for educational technology platforms that must balance operational costs against assessment validity. This work represents the first application of online control mechanisms to adaptively select prompting strategies in AES, transforming prompt selection from an offline hyperparameter optimization problem into an efficient online learning task.

\end{abstract}


\begin{table*}[!t]
\centering
\caption{Detailed Resource Consumption: MAB vs. Grid Search}
\begin{tabular}{llrr}
\hline
\textbf{Method} & \textbf{Calculation} & \textbf{LLM Calls} & \textbf{Tokens} \\
\hline
Grid Search & 787 essays × 4 recipes & 7,870 & 10,915,411 \\
& Multi-Step: 787 × 4 calls × 2 recipes = 6,296 & & \\
& Single-Step: 787 × 1 call × 2 recipes = 1,574 & & \\
& Total: 6,296 + 1,574 = 7,870 & & \\
\hline
MAB & 500 total essays (each assigned to one of 4 recipes) & 1,697 & 2,964,444 \\
& Multi-Step+Ex: 332 × 4 calls = 1,328 & & \\
& Multi-Step-NoEx: 67 × 4 calls = 268 & & \\
& Single-Step+Ex: 56 × 1 call = 56 & & \\
& Single-Step-NoEx: 45 × 1 call = 45 & & \\
\hline
Reduction & & 78.4\% & 72.8\% \\
\hline
\end{tabular}
\label{tab:detailed_resource_comparison}
\end{table*}

\section{Introduction}
Recent research has shown that Large Language Models (LLMs) have demonstrated significant capabilities in automated essay scoring (AES) when using well-designed prompts. Various prompting techniques for AES have been investigated, including rubric decomposition \cite{lee2024unleashing}, few-shot approaches \cite{yoshida2025we}, chain-of-thought reasoning \cite{chu2024rationale}, and comparative judgment methods \cite{kim2024gpt}. Despite these advances, contemporary approaches typically employ fixed prompt selection; researchers either predetermine a template or conduct exhaustive evaluations on a restricted set of options before adopting a standardized solution \cite{stahl2024exploring}. 
This static approach fails to address two key practical limitations in real-world assessment scenarios: \begin{enumerate}
    \item Inference costs are heavily influenced by prompt length and call frequency \cite{mcdonald2024can}. Rich prompting strategies that incorporate few-shot examples or detailed rationales can consume more tokens than minimalist instructions, which can significantly impact operational costs.
    \item The optimal prompt configuration is not static; it evolves as models are updated, pricing structures change, or essay characteristics shift over time. In high-stakes educational settings, such as international language testing programs or large-scale MOOC platforms, applying every essay to every possible prompt configuration is financially unsustainable.
\end{enumerate}
We propose a novel, cost-aware framework that treats each prompt-model combination as an arm in a multi-armed bandit (MAB) controller. During the scoring process, the agent observes a reward signal based on the agreement between the LLM's predicted score and the examiner's ground-truth score. This enables the system to incrementally concentrate calls on the most reliable and cost-effective prompt configurations. This approach transforms prompt selection from an offline hyperparameter optimization problem into an online learning task, similar to recent MAB optimizers for prompt engineering \cite{shi2024best} and retrieval-augmented generation \cite{fu2024autorag}.

Our preliminary experiments on the IELTS Writing Task 2 corpus from the IELTS Writing Scored Essays Dataset \cite{ielts2024dataset} demonstrate the effectiveness of the framework. The bandit-based approach achieves comparable Quadratic Weighted Kappa (QWK) scores to exhaustive grid search while requiring approximately one-tenth of the LLM calls, showing significant potential for reducing operational costs without compromising scoring quality.

This work makes two key contributions. First, to our knowledge, it represents the first application of online control mechanisms, whether bandit-based, reinforcement learning, or otherwise, to select prompting strategies in AES in an adaptive manner. All prior published work has relied on fixed prompt templates. Second, by tracking token usage and latency alongside agreement metrics, we produce the first cost-reliability learning curves for essay scoring. These findings offer actionable insights for test providers and educational technology platforms that must strike a balance between cost considerations and psychometric validity.

This paper reports work in progress. To date, we have implemented the bandit controller with four prompting approaches (single-step with examples, single-step without examples, multi-step with examples, and multi-step without examples). The four approaches are described in Section 3. For this initial stage of our study, we have used Google Gemini Flash 2.5. In our future steps, we will extend the study to additional models (GPT-4, Llama-3 70B) and the ASAP dataset \cite{asap2024dataset} to assess generalizability. By releasing our code and policy logs, we aim to stimulate further research into adaptive, cost-efficient LLM-based grading in educational assessment.

\section{Related Work}

The related work review is organized into three parts. We begin with LLM-based essay-scoring studies, summarizing research on zero-shot prompting, rubric-aligned, few-shot, and chain-of-thought methods, and noting their reliance on static, cost-blind grid searches. Next, we move to the writing domain to survey adaptive prompt and model-selection techniques in NLP, focusing on multi-armed bandit and other online controllers that optimize accuracy–cost trade-offs for retrieval and question-answering tasks. Finally, we summarize the public essay datasets that underpin most evaluations and pinpoint the still-unfilled gap: no prior work combines these adaptive controllers with essay scoring. This structure clarifies how our study builds directly on advances in prompt engineering and presents the first cost-aware adaptive framework for automated essay scoring.

\subsection{LLM-Based Automated Essay Scoring and Prompting Strategies}

Large language models (LLMs) have recently demonstrated effectiveness in grading student writing, thereby eliminating the need for task-specific fine-tuning. Lee et al.'s Multi-Trait Specialization (MTS) framework shows that a zero-shot, rubric-decomposed prompt can lift GPT-3.5 and Llama-2-13B to state-of-the-art Quadratic-Weighted-Kappa (QWK) on the ASAP and TOEFL11 benchmarks, outperforming a straightforward prompting instruction by up to 0.35 QWK \cite{lee2024unleashing}.

Stahl et al. \cite{stahl2024exploring} compare zero-shot, one-shot, and few-shot prompts (with and without chain-of-thought) for joint scoring and feedback generation, finding that AES accuracy improves modestly when the model is asked to explain its scores, but at the cost of much longer prompts.

Alternative prompting paradigms include comparative judgement: Kim \& Jo \cite{kim2024gpt} report that GPT-4, asked repeatedly to pick the better of two essays, surpasses a rubric-based direct-scoring prompt on the IELTS and ASAP sets \cite{asap2024dataset}.

For multi-trait scoring, Chu et al. \cite{chu2024rationale} generate trait-wise rationales with GPT-4 and feed them to a smaller student model; the rationale-augmented scorer beats strong baselines on ASAP++ \cite{mathias-bhattacharyya-2018-asap}and Feedback-Prize \cite{feedbackprize2021} datasets while offering transparent explanations.

\subsection{Cost-efficiency}
Most LLM-AES studies evaluate a small, fixed set of prompts offline and then deploy the single best template. Token budgets are rarely reported, even though few-shot prompts, including rubric and rationale, can exceed the context window of mid-tier LLMs. A recent study on rubric granularity by Yoshida \cite{yoshida2025we} shows that a simplified rubric maintains accuracy for three of four LLMs while cutting prompt length by more than half, underscoring the need for cost-sensitive experimentation. Nevertheless, the dominant evaluation paradigm remains static grid search: every prompt or configuration of prompts is tried on every essay, and the winner is chosen post-hoc.

\subsection{Adaptive Prompt and Model Selection in NLP}
Outside essay scoring, prompt and model choice have been framed as online decision problems. A framework TRIPLE, proposed by Shi et al. \cite{shi2024best} connects prompt optimization to fixed-budget best-arm identification and shows that a multi-armed-bandit (MAB) policy can identify a high-performing prompt on several NLP benchmarks while using only 50--80\% of the LLM calls required by exhaustive search.

In retrieval-augmented generation, AutoRAG-HP \cite{fu2024autorag} formulates hyperparameter tuning (e.g., k retrieved documents, prompt-compression ratio) as a hierarchical MAB; it matches grid-search recall with roughly 20\% of the API queries.

Somerstep et al.\cite{somerstep2025carrot} introduced CARROT (Cost AwaRe Rate Optimal Router), which applies a contextual bandit router to select, for each query, the cheapest LLM that still meets a target quality level, yielding substantial cost savings without compromising quality.

These successes demonstrate that adaptive controllers can maintain task performance while reducing token budgets; yet, none of these techniques has been incorporated into automated essay scoring. Our work addresses that gap by embedding an MAB inside an AES pipeline to select among four prompting techniques on the fly, delivering human-level reliability with a significant reduction in tokens.

\subsection{Public Essay-Scoring Datasets and the Remaining Gap}
LLM-AES research is typically benchmarked on the ASAP corpus (eight prompts, $\approx$ 12k essays) \cite{asap2024dataset}, the TOEFL11 corpus of 12,100 non-native essays \cite{toefl2013dataset}, and the newer IELTS Writing Band-Score set ($\approx$ 1200 essays) \cite{ielts2024dataset}.

All published LLM studies apply fixed prompt templates to these datasets; none employ bandits or any online search to allocate prompt-model calls. Consequently, our study is the first to embed a bandit controller inside an LLM-based AES system, adaptively routing each essay to the most cost-effective prompting strategy and thereby uniting efficiency research in NLP with high-stakes writing assessment.

\section{Methodology}

This study employs a novel approach to automated IELTS essay grading using LLMs with MAB optimization. We divided the IELTS dataset \cite{ielts2024dataset} into Task 1 and Task 2 subsets, focusing exclusively on Task 2 essays for this analysis. We developed a modular system with four distinct grading "recipes": multi-step criteria-based assessment with and without essay examples, and single-step direct scoring with and without examples.  The multi-step approach separately evaluates Task Response, Coherence and Cohesion, Lexical Resource, and Grammatical Range and Accuracy (official IELTS key assessment criteria \cite{IELTS_Writing_Criteria}) before calculating an overall score, while single-step methods directly predict the overall band score. 
For optimization, we implemented an epsilon-greedy MAB algorithm that balances exploration and exploitation to identify the most effective grading strategy. The reward function incorporates both prediction accuracy (measured by the negative absolute error against human scores) and optionally penalizes token usage for efficiency. We also conducted exhaustive grid search evaluations as a baseline comparison.
All experiments utilized Google's Gemini 2.5 model, selected for its balance of performance accuracy, low latency, and cost-effective API access. Experiments were performed on the Task 2 essay subset with human-assigned scores, employing concurrent processing to maximize throughput. Performance metrics included Mean Absolute Error (MAE) against human scores, token utilization, latency, and estimated API costs, allowing for a comprehensive assessment of each recipe's effectiveness and efficiency. The complete implementation code is publicly accessible on GitHub \cite{manakina2025_aesagent}.

\Needspace{6\baselineskip} 
\subsection{Dataset}
For our experiments, we utilized the IELTS Writing Scored Essays Dataset, a publicly available resource on Kaggle. This collection comprises 787 Academic Task 2 compositions, each accompanied by an official band score ranging from 1 to 9, assigned by qualified IELTS examiners. The compositions respond to conventional Task 2 instructions that challenge candidates to develop arguments or present perspectives on various societal and academic subjects. This corpus serves as an ideal testing ground for evaluating automated grading systems within the context of authentic, consequential language assessment.

\subsection{Dynamic Prompt Assembly}
Our system employs a sophisticated two-tiered approach to prompt engineering. The core architectural design separates prompt templates from their dynamic instantiation, enabling systematic experimentation with different prompting strategies.
The foundation of our approach is the prompt library module, which serves as a centralized repository of templated prompts. It contains detailed system prompts that establish the LLM's role as an IELTS examiner, along with specific instructions for each assessment criterion. For calibration purposes, some templates incorporate annotated sample essays (high and low-scoring) to provide reference points for the model's evaluation. The library maintains distinct template variants for each experimental condition (multi-step vs. single-step assessment, with vs. without examples).
At runtime, the essay grader agent dynamically assembles the final prompts through the following process:
\begin{enumerate}
\item Selection of appropriate system and user templates based on the current grading recipe
\item Injection of the specific essay question and text into the templates via string formatting
\item Construction of properly structured message dictionaries with distinct ``system'' and ``user'' roles
\item Assembly of these messages into the final prompt sequence transmitted to the LLM
\end{enumerate}
This modular design allows us to systematically compare different prompt structures while maintaining consistent content across experimental conditions. For multi-step assessment recipes, the system sequentially generates criterion-specific prompts for Task Response, Coherence \& Cohesion, Lexical Resource, and Grammatical Range \& Accuracy, before programmatically calculating the overall score. In contrast, single-step recipes construct comprehensive prompts requesting direct overall assessment, with or without calibrating examples.

\section{Results}

\subsection{Learning Behavior and Approach Selection}
The Multi-Armed Bandit (MAB) algorithm demonstrated clear preferences among the four grading recipes throughout our experiments. Figure~\ref{fig:mab_avg_reward} illustrates the cumulative average shaped reward for each approach over 500 steps. After an initial exploration phase with considerable fluctuation, the algorithm's assessment stabilized around step 100. The multi-step approach with calibration examples (Multi-step Ex) consistently achieved the highest reward values (approximately -0.75), followed by the single-step approach with examples (Single-step Ex) at around -1.0. 

\begin{strip}
  \centering
  \vspace{0.3 cm}
  \includegraphics[width=\textwidth]{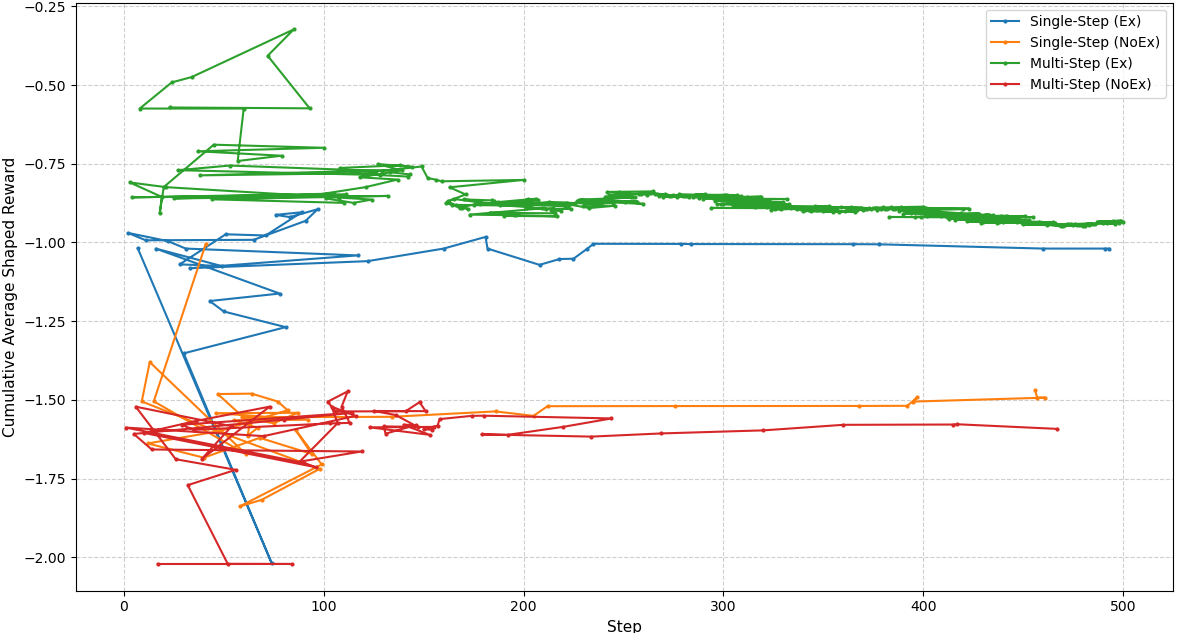}
  \captionof{figure}{Cumulative average shaped reward over steps}
  \label{fig:mab_avg_reward}
\end{strip}

\begin{figure}[t]
  \centering
  \includegraphics[width=\columnwidth]{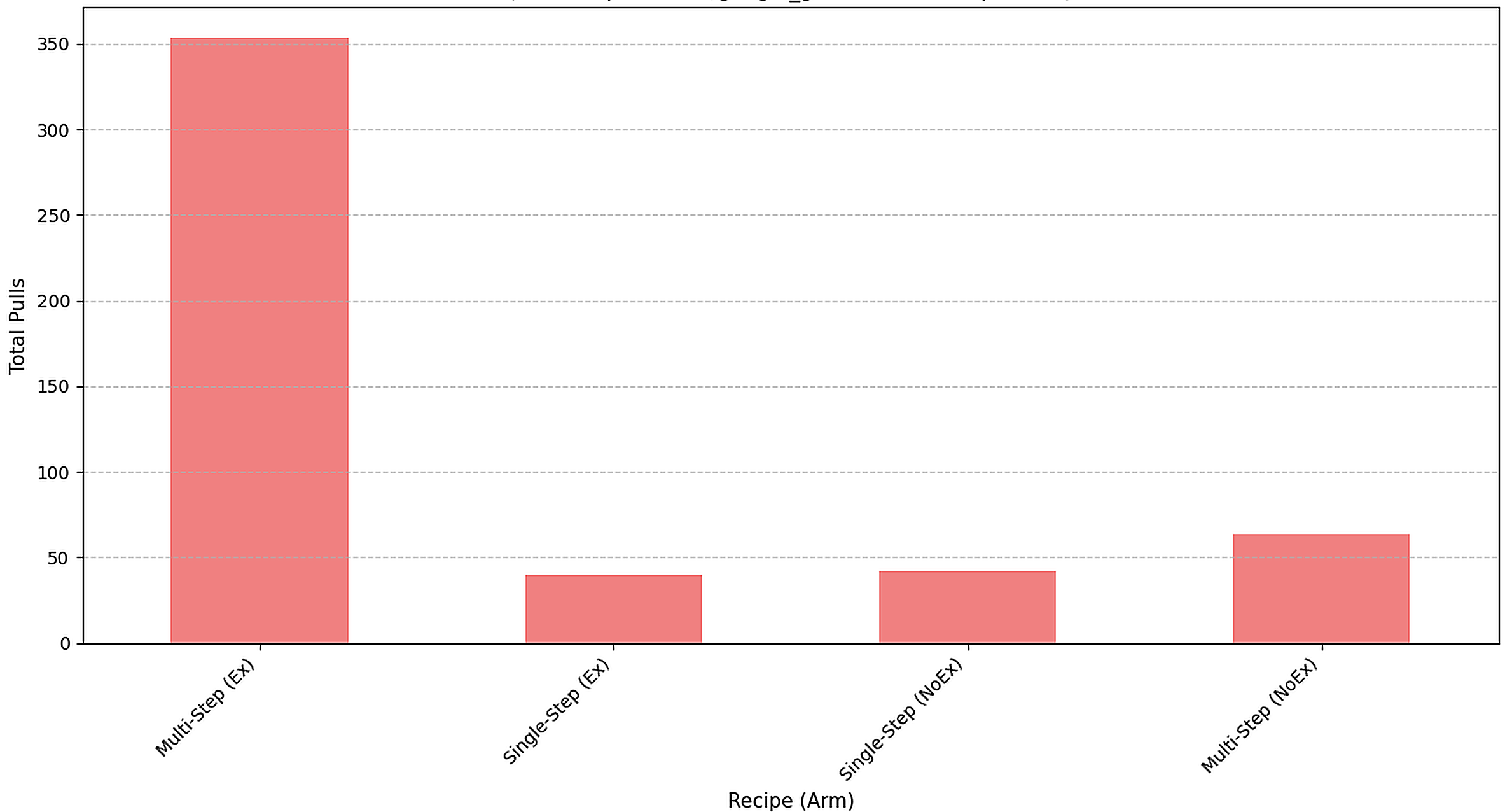}
  \caption{Distribution of arm pulls across the four grading recipes during the MAB experiment. The histogram shows the total number of times each recipe was selected by the algorithm, reflecting its learned preferences.}
  \label{fig:arm_pulls}
\end{figure}

The approaches without examples performed notably worse, with Single-step NoEx at approximately -1.5 and Multi-step NoEx showing the lowest rewards at about -1.6.
This learning behavior directly influenced arm selection frequency, as depicted in Figure~\ref{fig:arm_pulls}. 
The MAB algorithm favored the Multi-step Ex recipe, allocating approximately 350 pulls to this approach, over 70\% of the total experiment. The remaining three approaches received similar, but much lower attention, with each receiving between 40 and 65 pulls, demonstrating the algorithm's strong preference for the most effective recipe.

\subsection{Scoring Accuracy}
\begin{figure}[t]
  \centering
  \includegraphics[width=\columnwidth]{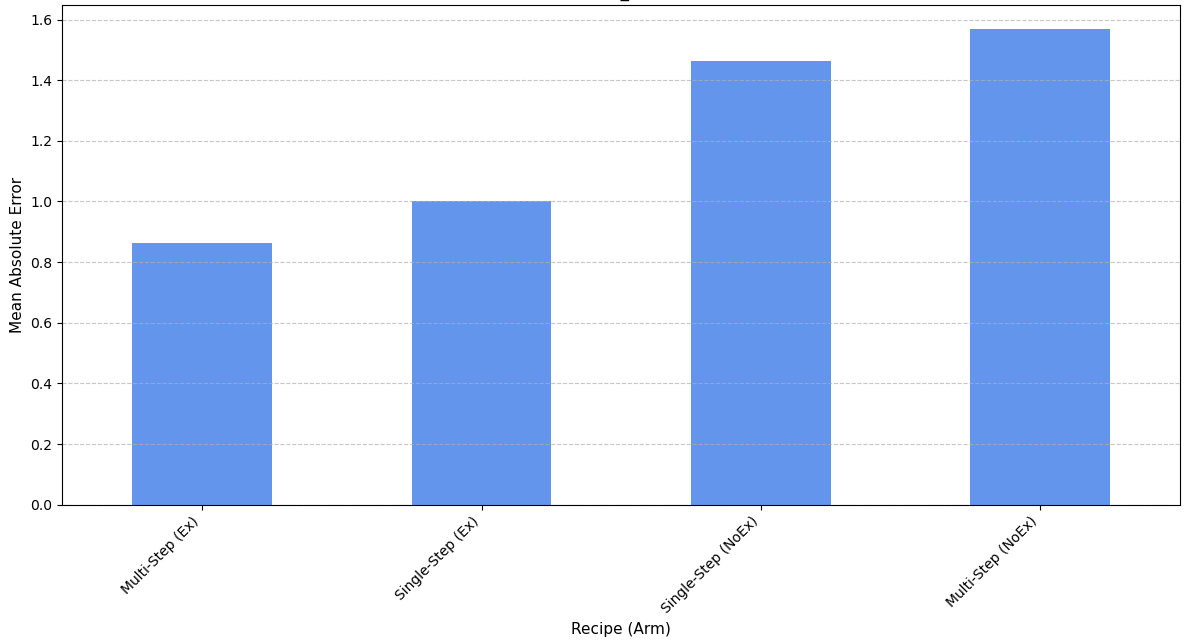}
  \caption{Mean Absolute Error (MAE) between predicted and human-assigned scores for each grading recipe in the MAB experiment. Lower values indicate greater scoring accuracy}
  \label{fig:mae}
\end{figure}

The MAB's preference for the Multi-step Ex approach is justified by its superior accuracy metrics. Figure~\ref{fig:mae} displays the Mean Absolute Error (MAE) for each recipe, revealing that Multi-step Ex achieved the lowest error rate (approximately 0.85), followed by Single-step Ex (1.0). The approaches without examples performed substantially worse, with Single-step NoEx showing an MAE of approximately 1.45 and Multi-step NoEx demonstrating the highest error rate at 1.55.

\begin{figure}[t] 
  \centering
  \includegraphics[width=\columnwidth]{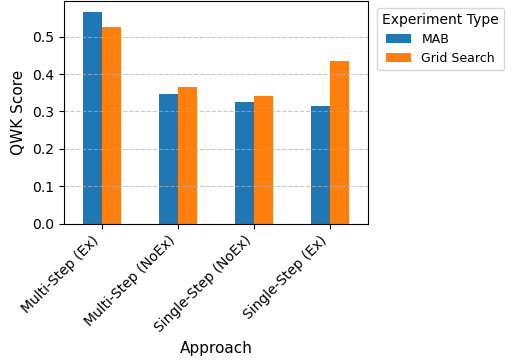} 
  \caption{Quadratic Weighted Kappa (QWK) scores comparing agreement with human graders for each recipe across both MAB and Grid Search experiments. Higher values indicate better agreement with human assessments. The comparison shows consistency between experimental approaches.}
  \label{fig:qwk}
\end{figure}

Our Quadratic Weighted Kappa (QWK) analysis in Figure~\ref{fig:qwk} corroborates these findings. QWK is a statistical measure of inter-rater reliability that accounts for both agreement and the magnitude of disagreement between ratings, making it particularly suitable for educational assessment tasks with ordinal scales. Unlike MAE, which treats all disagreements linearly, QWK penalizes larger disagreements more heavily, with values ranging from 0 (no agreement) to 1 (perfect agreement). The analysis shows that Multi-step Ex achieved the highest QWK score (approximately 0.55) across both MAB and Grid Search experiments, indicating the strongest agreement with human graders. Interestingly, Multi-step NoEx performed second-best in QWK (around 0.35), despite its poor MAE, suggesting that it occasionally captures assessment patterns that align with human judgment, despite the higher average error.

\subsection{Cost-Efficiency Tradeoffs}
\begin{figure}[t] 
  \centering
  \includegraphics[width=\columnwidth]{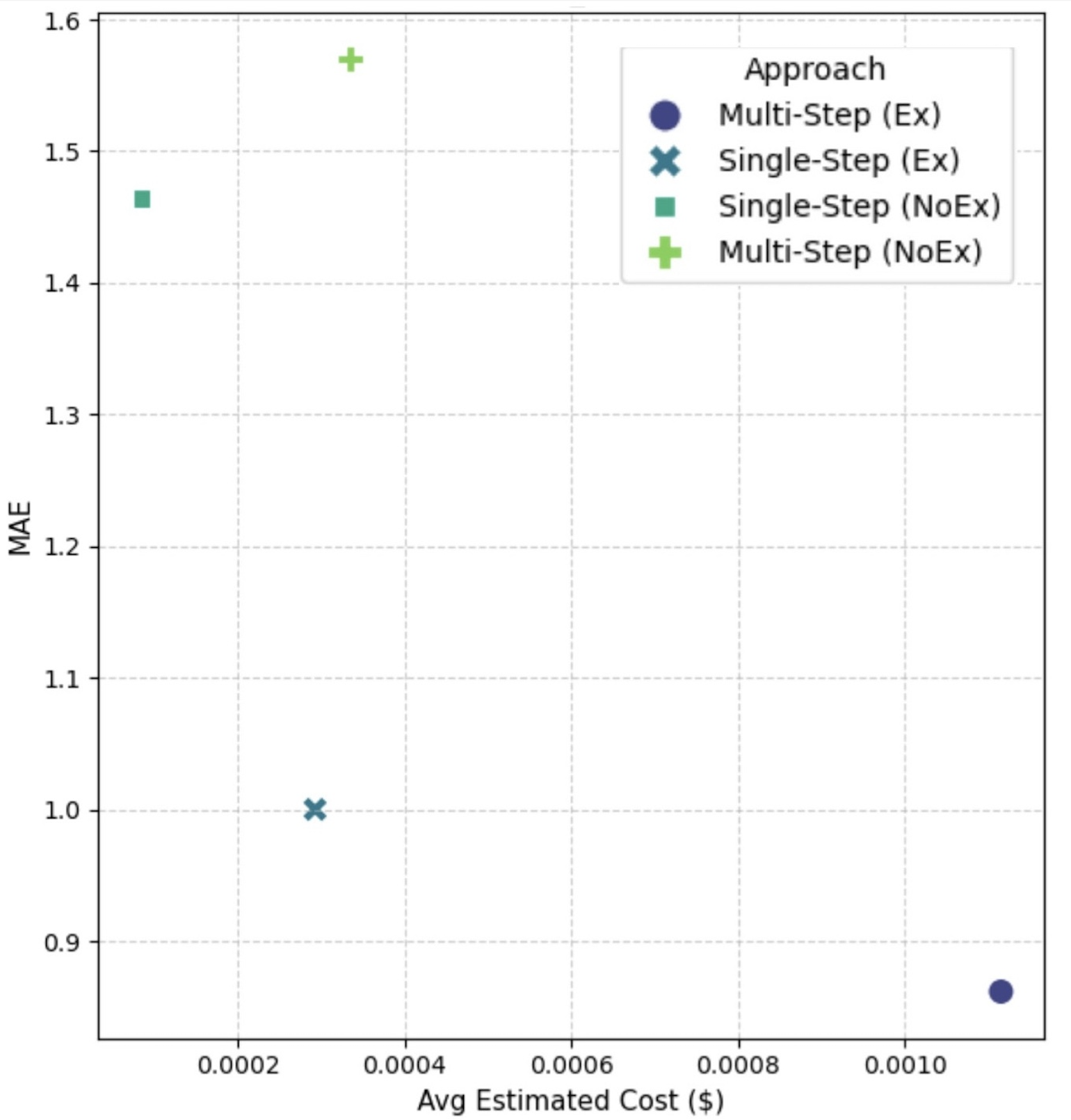} 
  \caption{Accuracy–cost tradeoff for four grading recipes. Multi-Step (Ex) achieves the lowest error (MAE 0.85) at a higher average API cost of about \$0.0011 per essay. Single-Step (Ex) provides a balanced option (MAE 1.0 at \$0.0003). Single-Step (NoEx) is the most economical (\$0.0002) but with higher error (MAE 1.45), while Multi-Step (NoEx) performs worst overall (MAE 1.55 at \$0.0004)}
  \label{fig:mae_vs_cost}
\end{figure}

While the Multi-step Ex approach demonstrated superior accuracy, our cost analysis revealed important efficiency implications across recipes. Figure~\ref{fig:mae_vs_cost} plots MAE against average estimated API cost per grading attempt, highlighting the accuracy-cost tradeoff. Multi-step Ex, while most accurate (MAE of 0.85), incurred a relatively high cost of approximately \$0.0011 per essay. Single-step Ex offered a balanced alternative with moderate accuracy (MAE of 1.0) at a lower cost (\$0.0003). The Single-step NoEx approach provided the most economical option at \$0.0002 but with higher error (MAE of 1.45), while Multi-step NoEx performed poorly on both metrics with the highest error (MAE of 1.55) at a moderate cost (\$0.0004).

\begin{figure}[!htbp]
  \centering
  \includegraphics[width=\linewidth]{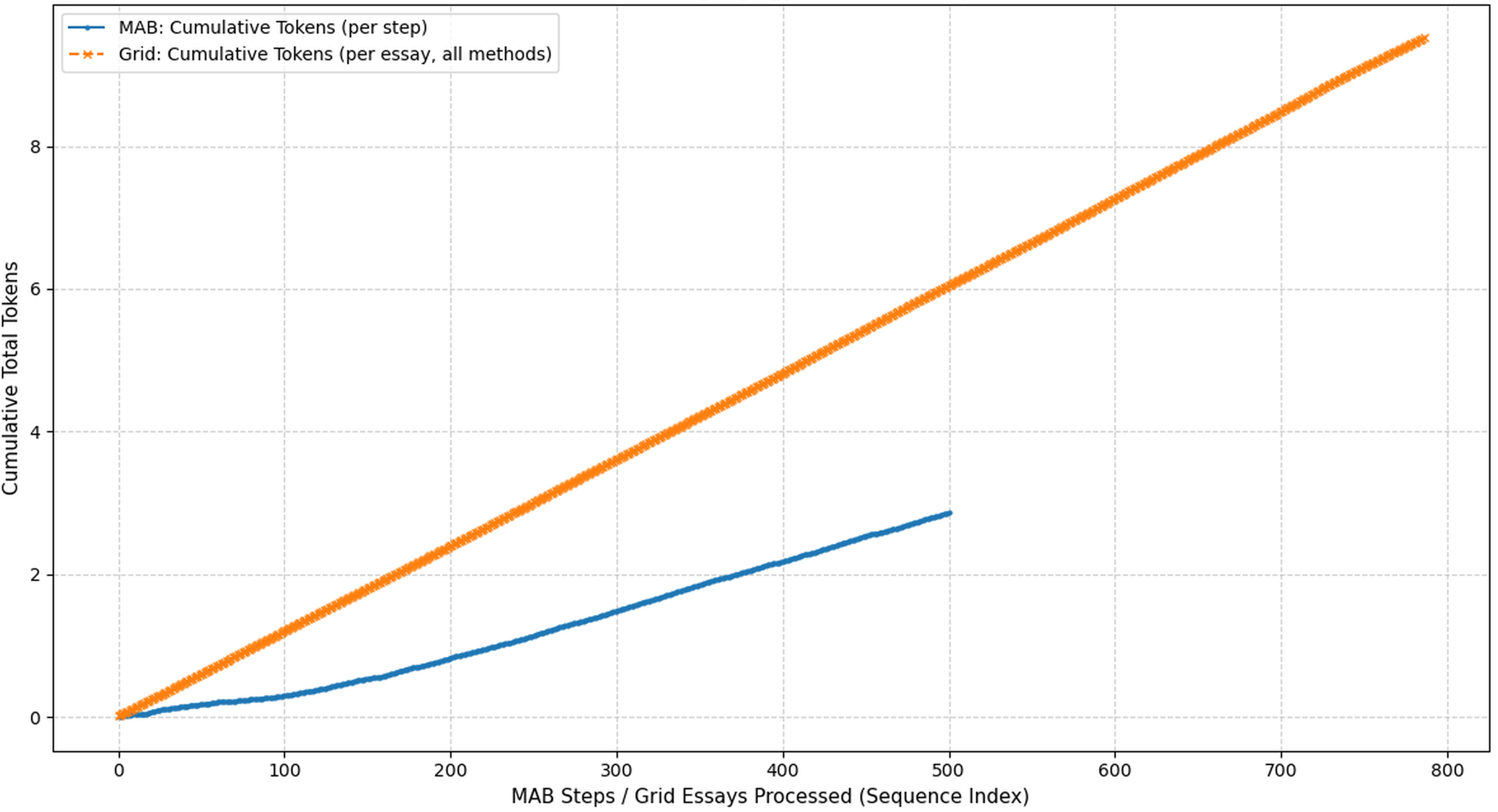}
  \caption{Cumulative token consumption over time for MAB versus Grid Search approaches. The steeper slope of the Grid Search curve demonstrates the efficiency advantage of the adaptive MAB strategy, which consumes approximately one-fourth the tokens for comparable coverage.}
  \label{fig:cumulative_tokens}
\end{figure}

 The cumulative token consumption comparison in Figure~\ref{fig:cumulative_tokens} demonstrates the significant efficiency advantage of MAB over Grid Search.

\begin{figure}
  \includegraphics[width=\linewidth]{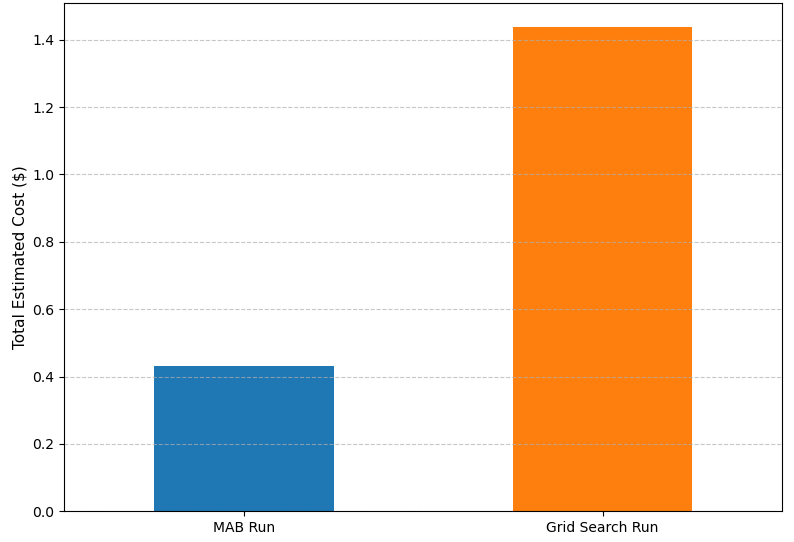}
  \caption{Total estimated API cost comparison between MAB and Grid Search experiments using Google's Gemini 2.5 model. The MAB approach demonstrates approximately 75\% cost reduction while maintaining comparable grading accuracy.}
  \label{fig:total_cost}
\end{figure}

While Grid Search exhaustively evaluates all approaches for each essay, consuming tokens at approximately five times the rate of MAB, the adaptive MAB strategy selectively allocates resources to promising approaches. At the experiment conclusion, Grid Search had consumed approximately 9 million tokens compared to MAB's 1.8 million for similar coverage. This efficiency translated directly to cost savings, as shown in Figure~\ref{fig:total_cost}, where MAB reduced total experimental costs by approximately 70\% compared to Grid Search (\$0.4 versus \$1.4) while maintaining comparable accuracy outcomes.

\subsection {Ablation Study}

To investigate the impact of prompt design on grading accuracy, we conducted an ablation study focusing on the inclusion of detailed explanations for assessment criteria. In all prompts for our primary experiments, we incorporated comprehensive descriptions of how each IELTS criterion is evaluated, stated in the official IELTS Writing Key Assessment Criteria document \cite{IELTS_Writing_Criteria}. For example, the Task Response assessment guidelines include detailed points such as: 1) How fully the candidate responds to the task; 2) How adequately the main ideas are extended and supported; 3) How relevant the candidate's ideas are to the task; 4) How clearly the candidate opens the discourse, establishes their position, and formulates conclusions; 5) How appropriate is the format of the response to the task.

Similar detailed explanations were provided for Coherence \& Cohesion, Lexical Resource, and Grammatical Range \& Accuracy criteria in their respective prompts. We then compared these results against a control experiment using simplified prompts without these detailed assessment guidelines.
Our ablation study revealed a surprising finding: removing detailed assessment criteria descriptions from prompts actually improved grading accuracy across all approaches. The Multi-Step with Examples recipe without explicit rubrics achieved an MAE of 0.862 and QWK of 0.566, outperforming its rubric-enhanced counterpart (MAE 0.965, QWK 0.485) while consuming fewer tokens (7,402 vs. 7,862) and reducing costs. This suggests that detailed rubric explanations may introduce noise or unnecessary constraints that interfere with the LLM's inherent understanding of essay quality. The MAB controller showed a stronger preference for the simplified approach, allocating 70.8\% of pulls to Multi-Step with Examples in the Basic experiment versus 66.4\% in the Detailed version. These results challenge the conventional wisdom that more detailed assessment guidelines necessarily lead to better automated scoring performance.

\section {Discussion \& Future work}
Our findings demonstrate that adaptive prompt selection via MAB can achieve substantial cost savings without compromising scoring quality, a finding that is particularly valuable for large-scale assessment programs. Surprisingly, our ablation study revealed that simplified prompts without detailed rubric explanations outperformed comprehensive ones, suggesting LLMs may have internalized academic writing assessment norms during pre-training. The strong performance of multi-step approaches aligns with human assessment practices and may apply to other LLM assessment applications.

This is a work-in-progress in paper; therefore, it has several limitations that we are planning to address in our next steps.  At this initial stage, we used only one LLM model (Gemini 2.5) exclusively on IELTS Task 2 essays, and our epsilon-greedy MAB implementation could be further refined.  We maintained a constant exploration rate ($\epsilon$ = 0.2) throughout our experiments to ensure sufficient exploration of all arms, although adaptive  $\epsilon$ strategies will be investigated in future work. Future work will also explore contextual bandits that incorporate essay-specific features, evaluate performance across diverse models and essay types, develop dynamic cost models that adapt to changing pricing structures, and investigate hybrid approaches that combine single-step efficiency with multi-step accuracy when needed.

\section{Conclusion}

This paper introduced a MAB-based approach to automated essay scoring that adaptively selects optimal prompting strategies. Our initial experiments demonstrated comparable accuracy to exhaustive grid search while reducing LLM calls by 78.4\% and token consumption by 72.8\% 
 (Table 1). Multi-step assessment with calibration examples consistently performed best (MAE 0.86, QWK 0.57), while, surprisingly, simplified prompts outperformed those with detailed rubric criteria. By framing prompt selection as an online learning problem rather than static hyperparameter optimization, our approach enables assessment systems to continuously adapt to changing model capabilities, pricing structures, and essay characteristics. As LLMs integrate into high-stakes assessment, such frameworks that balance accuracy, cost, and adaptability will become increasingly valuable.

\bibliographystyle{abbrv}
\bibliography{references}

\balancecolumns

\end{document}